# Deep Learning Approach for Real-Time Contraction Detection in Laboring Dairy Cattle

Sebastian Puchala[1], Dominik Gronkiewicz[1], Marcin Gołębiewski[2], Grzegorz Grodkowski[2], Grzegorz Gwardys[1,3]

1 Promity, Warsaw, Poland
2 Institute of Animal Science, Warsaw University of Life Sciences, Warsaw, Poland
3 Faculty of Electronics and Information Technology, Warsaw University of Technology, Warsaw, Poland

## Abstract

Currently, the animal production sector is undergoing rapid development of techniques that allow automatic monitoring of animal behaviour and health. This paper presents a computer vision-based approach for the automatic recognition of labor contractions in cows. During the course of the study, videos depicting 100 cow births were collected. From the obtained videos, 8500 cow contraction actions and about 25 000 non-contraction sequences were extracted which constituted the dataset. Leveraging image processing techniques and deep learning algorithms like Convolutional Neural Networks (CNNs) and Vision Transformers (ViTs), our methodology extracts relevant features from video frames to detect and classify contractions. Our study demonstrates a high accuracy rate of over 95% in detecting labor contractions. This automated system has the potential to enhance productivity and animal welfare in the dairy industry by providing real-time monitoring and facilitating timely intervention during difficult calvings. The video demonstration of the system can be found at the following URL: https://www.youtube.com/watch?v=8fTSG5D7dZ0

## Introduction

Childbirth for both the cow and the calf is a critical period affecting the subsequent health of both the mother and the calfling [1-2]. Difficult births, known as dystocia, are common in dairy cows [10]. Studies show that dystocia rates in the United States range from 28.6 to 51.2% in primiparous cows and from 10.7 to 29.4% in multiparous cows [3-4]. Failure to detect difficult labor in a timely manner and implement birth assistance can result in an increased risk of stillbirth, they also increase the likelihood of birth complications and reduce the productivity and welfare of the cow and calf [1,2,5,7] which consequently reduces farm profit.

Predicting calving time facilitates labor organization, thus minimizing unnecessary pain and stress, especially in situations requiring intervention [6]. External signs such as pelvic ligament loosening, udder swelling, vulvar swelling and behavioral changes are often used to predict the onset of calving in dairy cows. However, visual observations are time-consuming. They require regular and continuous repetition around the clock. Therefore, systems have appeared on the market that allow automatic detection of the second phase of calving. These include the Moocall system consisting of a sensor placed on the tail. It measures contractions using an accelerometer. Another example of a sensor is the C6 Birth Control which requires sewing to the labia to function properly. It measures the degree of dilation of the genital tract. These types of birth detection systems are characterized by high accuracy and reliability, however, they require knowledge of how to install them and it is necessary to remove them after delivery. These devices can also be invasive to animals, especially if not installed properly. Infection can occur or, in the case of the Moocall device, tail necrosis.

An alternative to this type of sensor is to use camera images to analyze the behavior of calving cows. A solution like this eliminates the need for employees to observe the parturient cows and does not require the installation of any device on the animal's body. It is known that the cow's behavior before parturition changes significantly the time of lying in the lateral position increases [1], the time of raised tail increases [8], the time spent in foraging and chewing also decreases [9]. These changes are observed in the first and second stages of parturition, are an individual trait and can vary considerably between different breeds and calving rates [8], hence their correct interpretation is much more difficult and requires a lot of data. On the other hand, a common feature of all parturient cows is the occurrence of labor contractions in the second stage of labor, that is, the fetal displacement phase. In dystocia births, contractions are frequent and strong after which they weaken, a sign of the cow's weakness [5]. At this point, labor assistance is required.

The introduction of computer vision techniques provides a means to automate this recognition process, enabling real-time and continuous monitoring of labor contractions. By analyzing video streams captured from strategically placed cameras within calving areas, computer vision algorithms can extract relevant features to detect labor contractions accurately. Before the Deep Learning era, we could describe these features as motion vectors, texture information, or shape descriptors [10]. However today the burden of determining the appropriate features rests with neural networks and this is the approach we follow. Convolutional Neural Networks (CNNs) have demonstrated exceptional performance in image recognition tasks [11]., making them well-suited for the detection of labor contractions. By training CNN models on annotated datasets comprising synchronized video recordings and ground truth annotations of labor contractions, the algorithm can learn to identify contraction patterns. Additionally, Vision Transformers (ViTs) [12], a recent advancement in deep learning, offer an alternative approach for feature extraction and recognition, leveraging self-attention mechanisms to capture spatial relationships in video data.

While the specific domain of labor contraction recognition in cows using computer vision techniques is relatively unexplored, there are various relevant studies that highlight the application of computer vision in various aspects of livestock management.

Authors in work [13] provide an overview of computer vision-based technologies used for cow lameness detection and discuss the advantages and disadvantages of different camera types, also pointing challenges that remain in terms of detection efficacy and practical implementation. Bezsonov Oleksandr et al. [14] present an intelligent system that integrates synchronized 3D cameras and neural network modules for breed recognition and live weight estimation of cows. In [15] Nguyen Chuong et al. propose a prototype that utilizes embedded sensors to monitor cow welfare on farms, ensuring quality control and compliance. In [16], an embedded imaging system is presented to assess dairy cow drinking behaviour in relation to heat stress. In [17] a framework is introduced that predicts the lactation curve for dairy cows using the cow's historical milk yield sequence and reproductive and health events. There are also many works that propose working computer vision components that are not linked directly to the agricultural application, for example, cattle pose estimation algorithms [18-19].

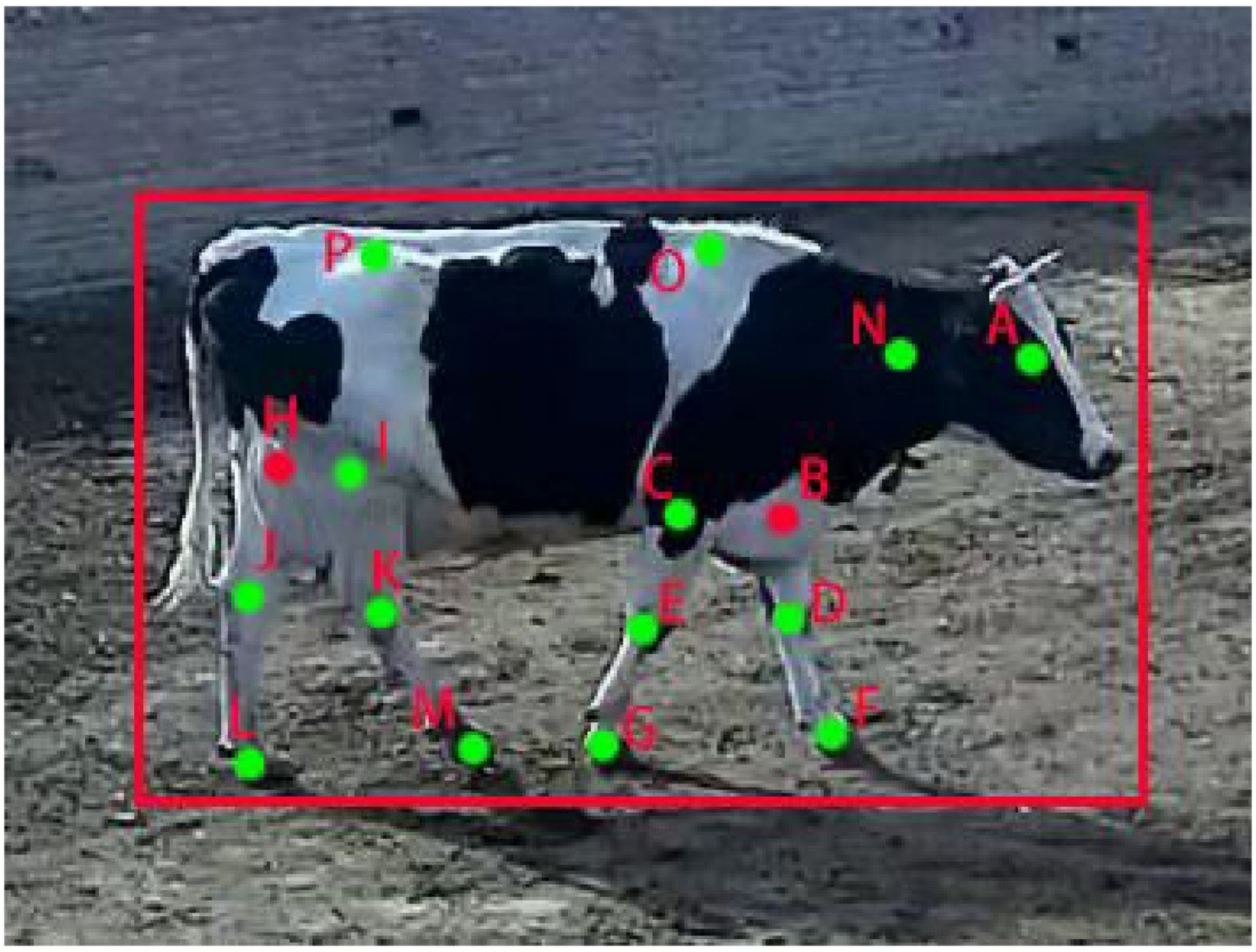

Fig 1. Example of a keypoint label template of the cow skeleton presented in [19].

By leveraging the capabilities of CNNs and ViTs, this study aims to develop an automated system capable of detecting labor contractions in cows. As stated before this topic is rather neglected, probably due to the cost of producing the relevant data set. We believe that bucketing contractions' amounts and setting thresholds is a good first step to notify famer about an impending birth – this conception is shown in figure 2 and at the following URL: https://www.youtube.com/watch?v=8fTSG5D7dZ0. Comparison between our study and similar works is conducted in the Discussion.

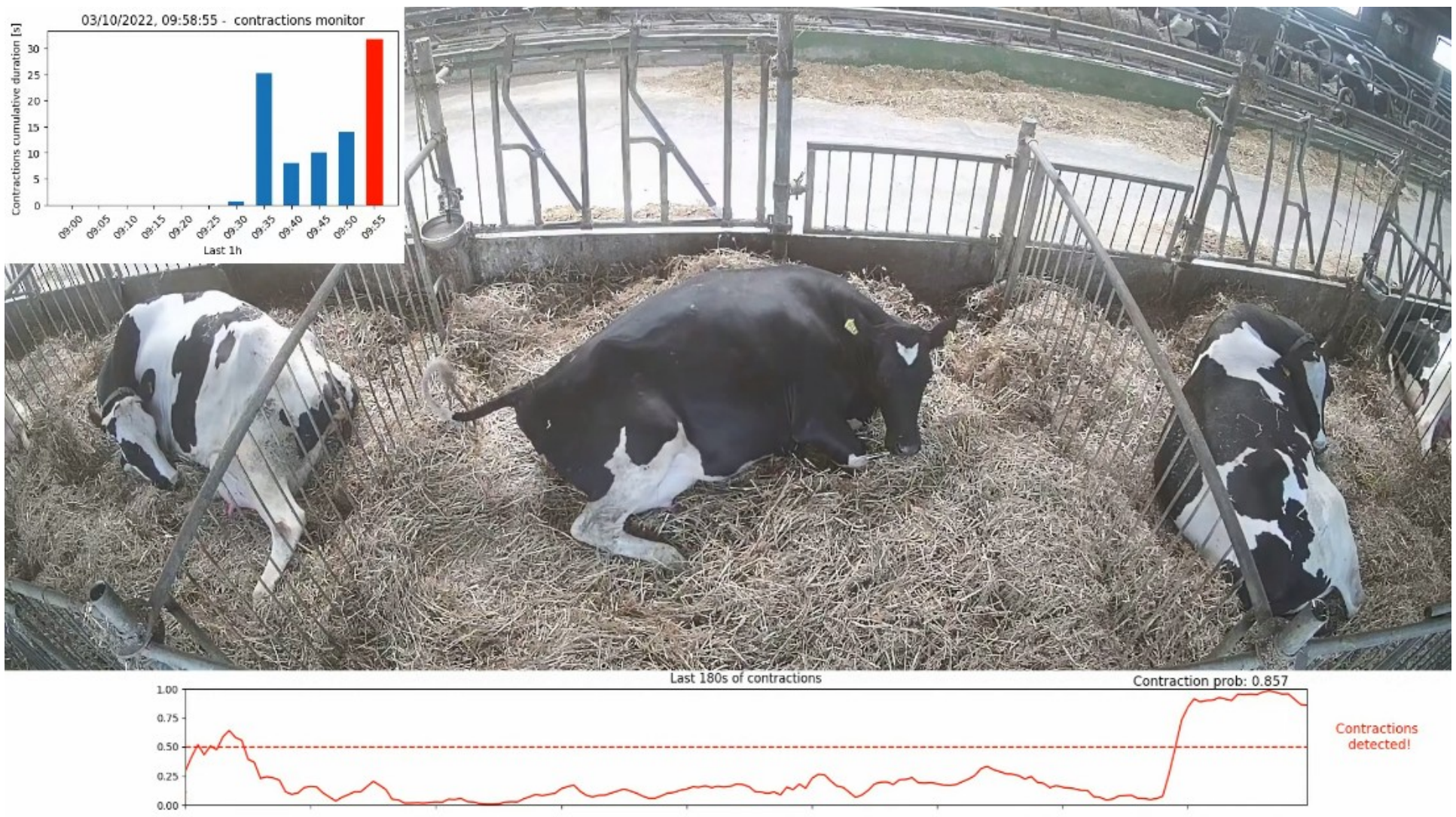

Fig 2. Screenshot of our demonstration. In the left top corner, there are amounts of contractions in the 5-minute buckets. The chart at the bottom of the photo shows the contractions' score at the current moment. In this example, the cow is during the contractions.

# Materials and Methods

The experiment was conducted on a high-input farm using 350 Polish Holstein-Friesian cows of the black-and-white type, fed on forage (TMR).
Prior to scheduled calving, cows were moved to individual farrowing stalls. Eight farrowing stalls were selected for the experiment. Each was equipped with 3 RGB infrared cameras arranged in such a way that each camera observed the cubicle from a different perspective. Data were collected from 1.10.2021 to 31.09.2022. During the experiment, 100 parturient cows were recorded. The data was acquired by the monitoring cameras, whose setup is described in the next section. During the experiment, neither the data acquisition equipment nor the experimental personnel contacted the cows and had no stress response to cows. Compared with traditional manual inspection and using wearable sensors, it can realize noncontact animal behavior detection and improve animal welfare. For all these reasons, there are no ethical issues.

## Data Acquisition

The availability of big and high-quality datasets is crucial for the development of any machine learning system. We have created a unique data acquisition center monitoring cow behavior using a variety of cameras. Recordings of cattle were taken in the following setup situated in the calving hall: 24 RGB infrared cameras, 4 rotating cameras, 2 fish-eye cameras, and 2 RGB+D cameras. In this study we concentrate only on data acquired from 24 RGB infrared cameras and the use of other data sources is the subject of the next works. These 24 cameras were placed in eight farrowing boxes (3 for each). The height of the camera suspension in farrowing boxes is 2-2.5m and size of each box is around 3x3 m.

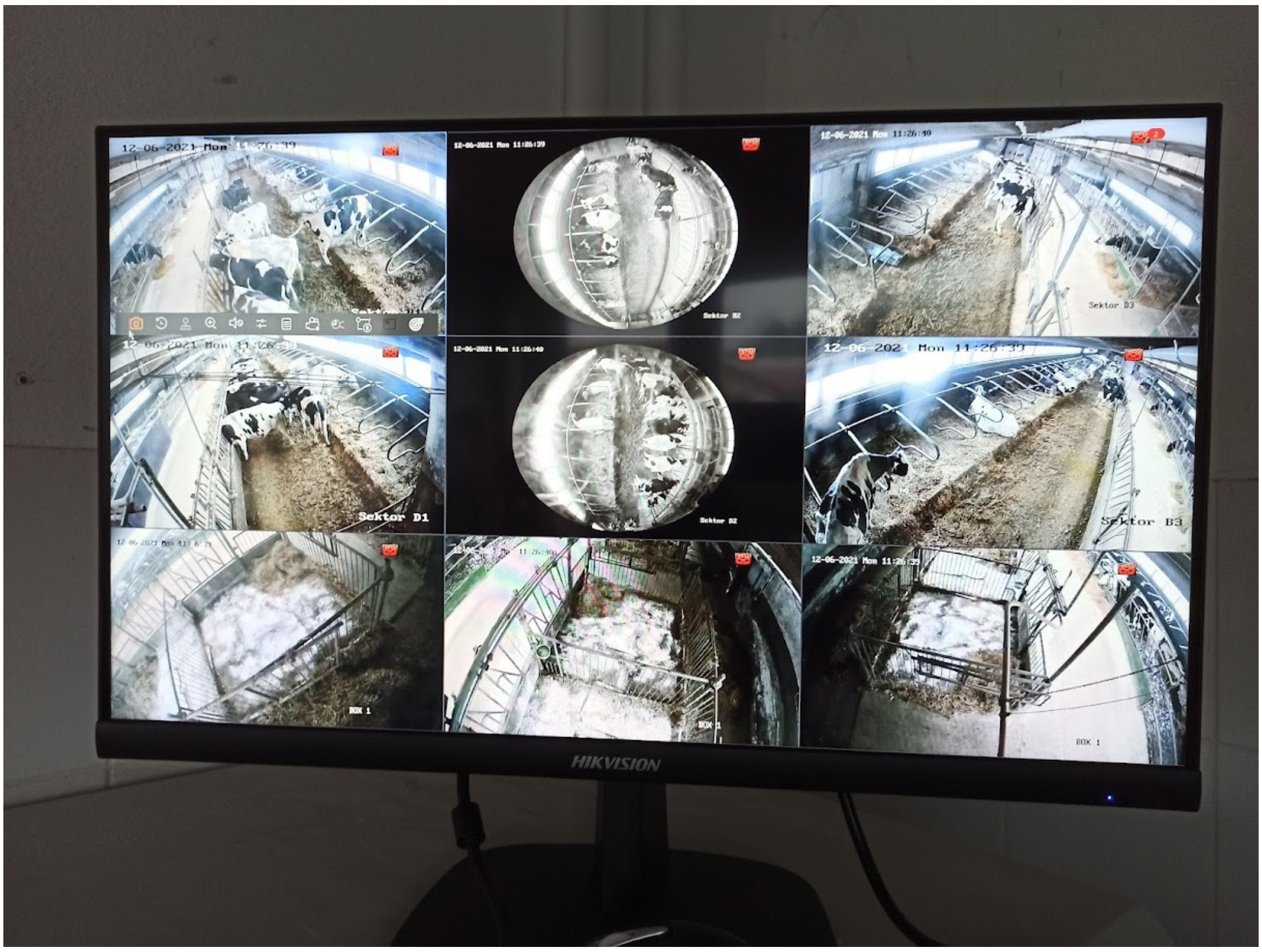

Fig 3. Photograph of a screen that presents views from various installed cameras. Visible ‘circle views’ are delivered by fish-eye cameras that were not used in this study.

All raw footage of both calving and daily cattle behaviour have been documented, creating vast and unique data collection. Using information about the time of calving, 4-hour recordings were extracted. Then, the moments of labor contractions were annotated using the CVAT software [26]. This enabled the creation of a specialized Contraction dataset. It contains about 8500 cow contraction actions and about 25 000 non-contraction sequences for 100 unique cows. Each action was recorded in a separate video lasting between 2 and 4 seconds. For each sample metadata like camera view (center, right, left), box number, or acquisition date were saved. This dataset is utilized to train a contraction detector since it contains information on the frequency of contractions prior to calving.

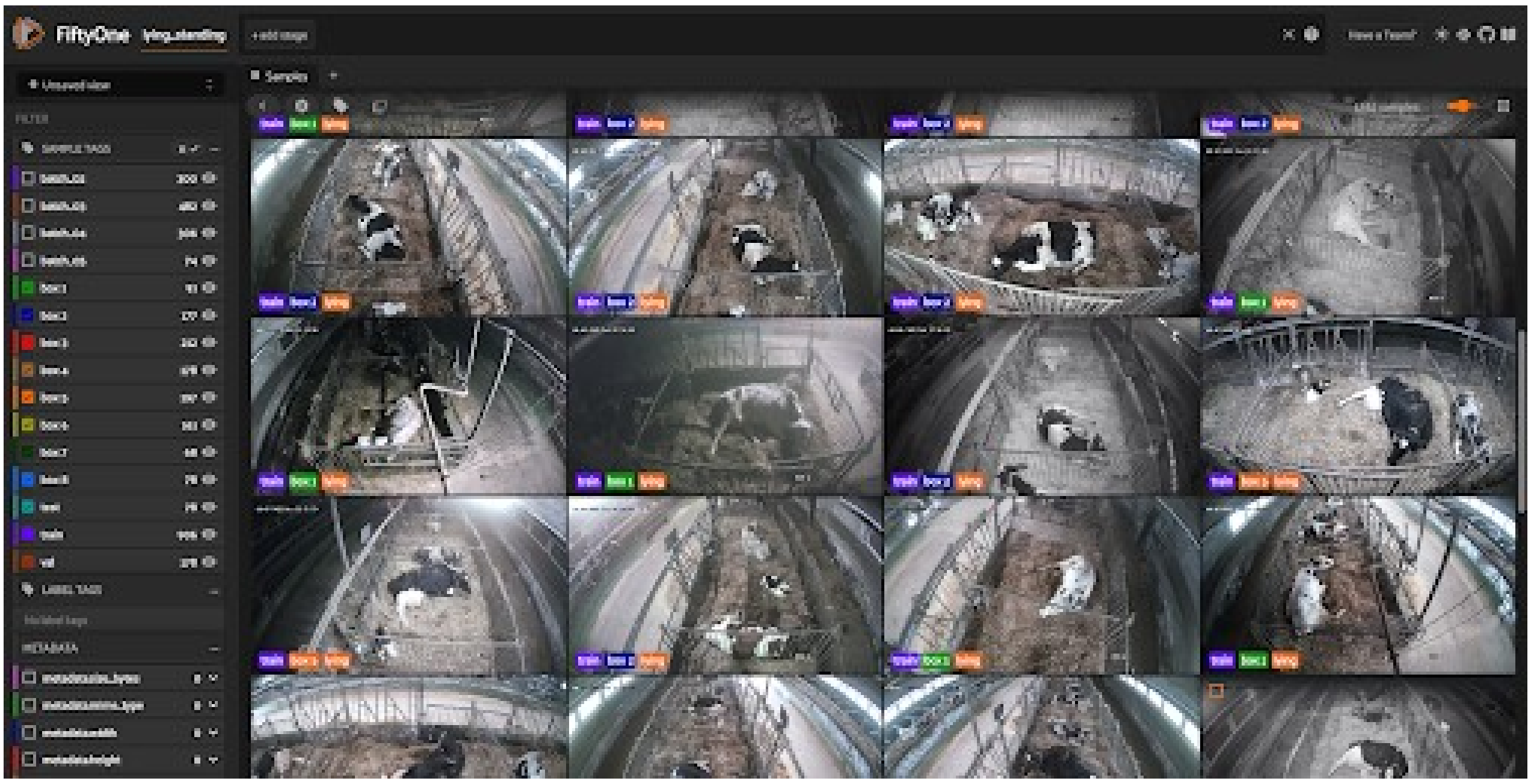

Fig 4. Screenshot of FiftyOne tool that presents various cows and allows to easily visualize errors – for example, restricting videos to False Positives for further investigation.

# Methods

While a binary classification using CNN is not a new concept, a data-centric approach was crucial to the success of our model. It is very important to set up training and test sets fairly to prevent data leakage and ensure the model is evaluating unseen data during the learning process.

We took a convenient approach by splitting the data based on the boxes the cows were in within the barn, which allowed us to evaluate the model on diverse data without overfitting on particular cows or camera angles. When using video instead of still images, it is necessary to select a representative sample of frames to use for training. For example, a 4-second video clip of a cow contraction at 30 frames per second would provide 120 frames, which is not practical for training a model due to the large amount of data involved. To overcome this problem, frame subsampling can be used to select a number of frames that are linearly spaced. This method results in a smaller volume of data for learning, such as an 8-frame clip with a volume of 8×3×height×width.

However, the temporal differences between frames must also be taken into account, as videos in the dataset may be of different lengths. This is where the sampling rate comes in, as it models the temporal differences between frames. By sampling clips with a fixed length, determined by the formula

, we can ensure that the model is trained on a representative sample of frames from each video. Both of these values, the number of subsampled frames (also called frame length) and the sampling rate, often presented next to model architecture as frame length x sampling rate, should be treated as hyperparameters of the model and tuned accordingly. Before training the model, it is essential to familiarize oneself with the dataset. While it's possible to review the data on disk, with a complex folder structure and annotations, it's more convenient to use specialized tools like e FiftyOne. With this tool, our team could quickly view all the data in a web app and filter it by tags and labels.

## Real-Time Contraction Detection Network

After conducting an extensive series of evaluations of various architectures, we decided to build our real-time contraction detection network around two models. We chose a Slow R50 model, which we tailored to our requirements from the SlowFast design [19], as well as an MViT [20] with a 32x3 setup.

### Customized Slow R50 Model

Our customized Slow R50 model is a modified version of the SlowFast [19] architecture. Part of the model's name, R50, results from the fact that the model processes video as a spatio-temporal volume, utilizing the ResNet-50 backbone's 3D convolutional layers. It contributes to capturing spatio-temporal features and thanks to utilizing residual connections (a key feature of the ResNet-50 architecture), mitigates the vanishing gradient problem and enhances gradient flow throughout the network.

SlowFast [19] architecture utilizes two distinct pathways to process video data. The slow pathway focuses on capturing spatial semantics by analyzing high resolution frames though at a lower frame rate. In contrast the fast pathway analyses low resolution frames at a higher frame rate to capture motion related information.

We considered the trade-off between computational complexity and model accuracy when reaching a decision, especially in terms of the capability to process data in real time on less powerful devices (like autonomous cameras).

In our specific application, we made an intriguing discovery – by exclusively utilizing the slow pathway of the SlowFast architecture, we were able to obtain satisfactory results in terms of both efficiency and accuracy for our task. This approach draws inspiration from an experiment mentioned but not fully explored in the original SlowFast paper [19], where the authors introduced a Slow-only model as a part of their ablation studies. Notably, even without incorporating the Fast pathway, they discovered that this variant maintained robust performance levels with reduced computational overhead. These observations strongly influenced our project's objectives and drove us to develop this idea into an effective, real-time solution for labor contraction detection in dairy cattle. Our work stands as one of the first implementations that build on and extend the initial Slow-only concept to practical applications in the field of animal welfare.

### Multiscale Vision Transformer (MViT)

The second model under consideration is the Multiscale Vision Transformer (MViT) architecture [20], formulated with a 32x3 configuration. MViT is a transformer-based architecture tailored for visual data processing, employing multiple channel-resolution 'scale' stages to establish a multiscale pyramid of feature activations. By doing so, MViT allows early layers to encode low-level visual information, while deeper layers focus on higher-level features to capture visual semantics effectively. MViT's design exploits the dense nature of visual signals and possesses a notable implicit temporal bias, rendering it highly advantageous for video models.

While it is true that the MViT model is more computationally intensive and slower than the Slow R50, it remains still viable for deployment on high-end embedded devices. We chose to explore this model to provide a slightly more robust, albeit heavier, option compared to the Slow R50, potentially offering superior accuracy in exchange for increased resource requirements. Our aim was to evaluate a range of models, balancing the trade-off between computational efficiency and prediction accuracy, to ultimately select the best fit for our specific needs.

## Training procedure

Both our customized Slow R50 and MViT models were initially pre-trained on the Kinetics dataset, an extensive collection of over 500,000 video clips across 600 human action classes. This initial training phase served as a solid foundation for fine-tuning the models on our dataset, leveraging the benefits of transfer learning. To train models we set learning rate of 0.0001. To handle multi-class classification tasks, we applied the cross-entropy loss function. Our optimizer of choice was the stochastic gradient descent (SGD), a standard in many deep learning applications. To manage the learning rate, we employed a reduce-on-plateau scheduler that adjusts the learning rate when the model's validation performance plateaus.

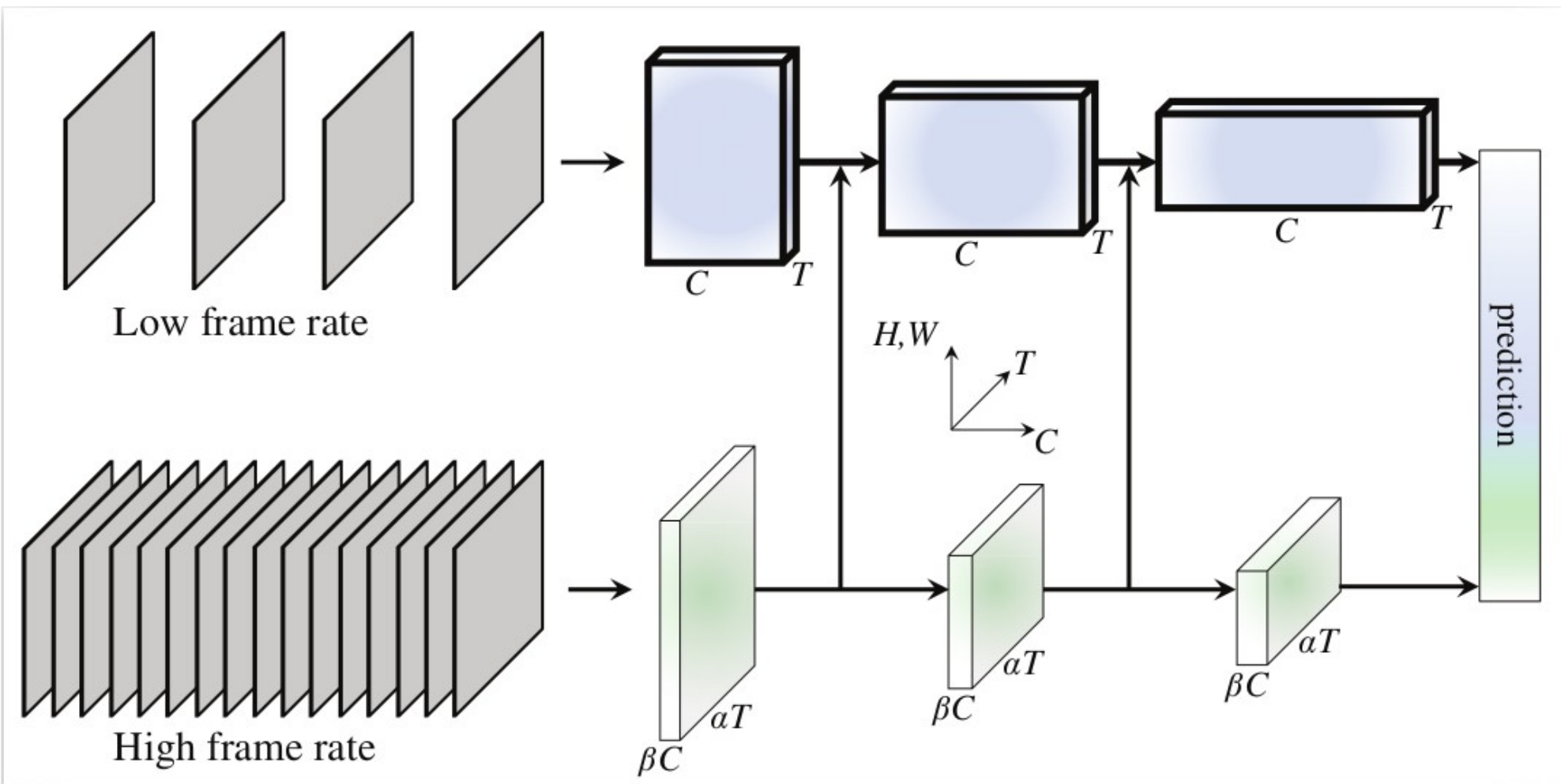

Fig 5. **A SlowFast network** with two pathways: Slow pathway characetrized by a low frame rate and low temporal resolution and Fast pathway with a high frame rate, ω × and higher temporal resolution. The graphic comes from [19].

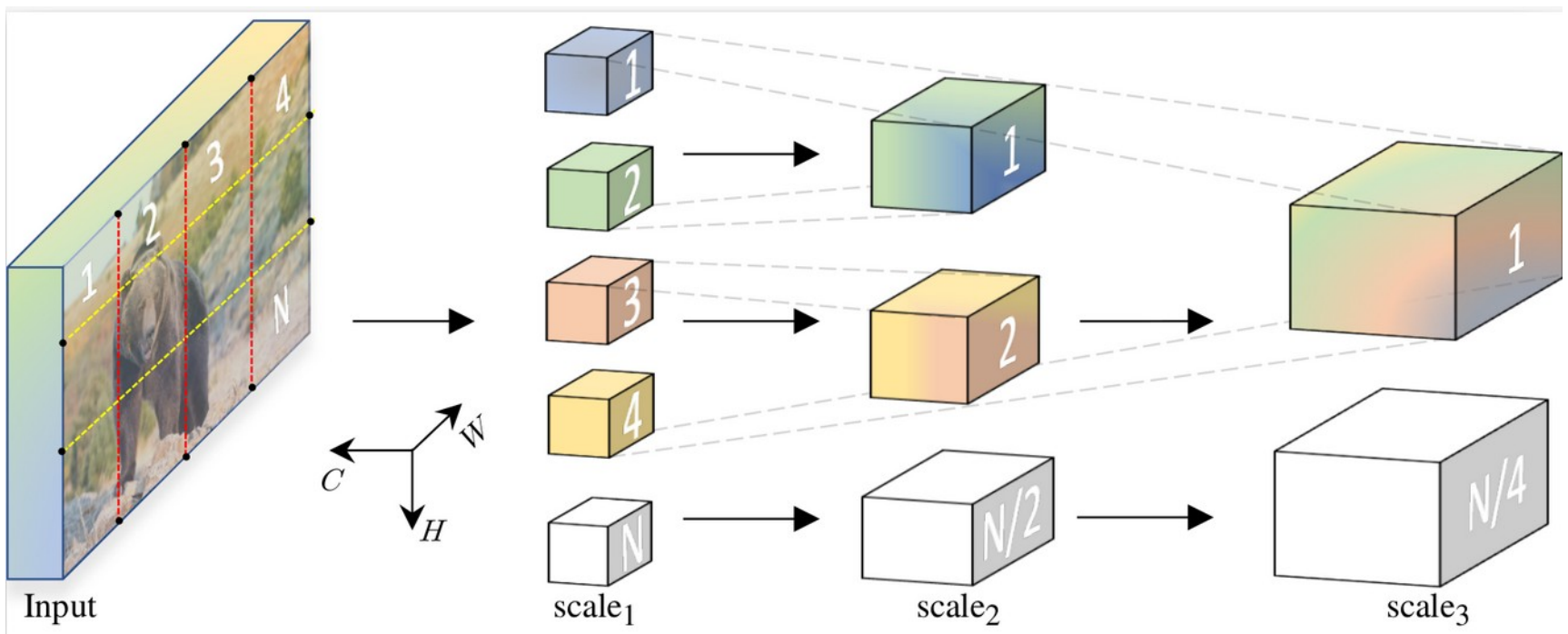

Fig 6. **Multiscale Vision Transformers** learn a hierarchy of coarse (in space) and complex (in channels) features. The graphic comes from [20].

# Results

The experiments were conducted on machine with AMD Ryzen 9 5950X 16-Core Processor and NVIDIA GeForce RTX 3090. To the main software components we can list CUDA (Version: 12.0), Python (Version: 3.10.6), Pytorch (Version: 1.12.1+cu116), Torchvision (Version: 0.13.1+cu116), Pytorch Video (Version: 0.1.5)

To ensure the robustness of our findings, we employed eight distinct boxes as experimental units, conducting eight experiments in which each box served as the exclusive test set, while the remaining seven boxes were utilized for model training purposes.

By systematically rotating the role of the test set across the eight experiments, we sought to capture a comprehensive representation of the dataset's variability and evaluate the models' efficacy. Subsequently, we analysed the statistical metrics derived from the outcomes of each experiment to ascertain the quality of the models. This evaluation framework enables us to draw conclusions regarding the detection of cattle contractions, enhancing the credibility of our research outcomes.

The findings of our study demonstrated that the Multiscale Vision Transformers (MViT) model exhibited superior performance compared to the alternative model investigated, as evidenced by its higher prediction accuracy. Nevertheless, it is important to acknowledge that both models yielded satisfactory results within the context of this research.

Of particular interest was the observation that the MViT model showcased a heightened resilience to data variations arising from the selection of specific experimental boxes. This characteristic suggests a greater degree of generalizability, indicating the MViT model's ability to maintain robust performance across diverse datasets. Such resilience to variations is a noteworthy attribute, underscoring the potential of the MViT model as a reliable and versatile tool for action classification tasks.

**Table 1. Evaluation of Slow R50 Model and MViT. Results are divided into boxes.**

| Type | Box 1 | Box 2 | Box 3 | Box 4 | Box 5 | Box 6 | Box 7 | Box 8 | Mean |
|---|---|---|---|---|---|---|---|---|---|
| **RGB_Slow_R50** | 90.01 | 84.04 | 88.36 | 81.71 | 88.99 | 88.38 | 92.46 | 81.76 | 86.96 ± 3.98 |
| **RGB_MViT** | 90.27 | 89.85 | 89.98 | 85.49 | 90.31 | 90.82 | 96.41 | 93.44 | 90.82 ± 3.13 |

Although graphical representations and statistical analyses are fundamental for assessing model performance, they may not provide precise insights into the underlying causes of mispredictions. In order to gain a deeper understanding of the model's limitations and areas of weakness, we conducted a comprehensive examination of the misclassified actions. In this regard, the FiftyOne app emerged as an indispensable resource, offering a valuable suite of features for data visualization and evaluation.

**Table 2. Evaluation of Slow R50 Model and MViT.**

| Type | Train Batches | Test Batches | Box 1 | Box 2 | Box 3 | Box 4 | Box 5 | Box 6 | Box 7 | Box 8 | Mean |
|---|---|---|---|---|---|---|---|---|---|---|---|
| **RGB_Slow_R50** | **1,2** | **1,2** | **90.01** | **84.04** | **88.36** | **81.71** | **88.99** | **88.38** | **92.46** | **81.76** | **86.96 ± 3.98** |
| **RGB_Slow_R50** | **1,2,3** | **1,2,3** | **92.35** | **85.53** | **90.14** | **87.81** | **89.55** | **89.99** | **89.51** | **90.58** | **89.43 ± 2.02** |
| **RGB_Slow_R50** | **1,2,3** | **1,2** | **92.29** | **79.04** | **89.33** | **83.6** | **90.4** | **90,07** | **95.4** | **88.8** | **88.62± 5.1** |
| **RGB_MViT** | **1,2** | **1,2** | **90.27** | **89.85** | **89.98** | **85.49** | **90.31** | **90.82** | **96.41** | **93.44** | **90.82± 3.13** |
| **RGB_MViT** | **1,2,3** | **1,2,3** | **92.62** | **92.47** | **92.34** | **91.36** | **91.28** | **93.74** | **93.08** | **94.93** | **92.72 ± 1.21** |

Consider the case of the Slow 50 model applied to box 4, which represented one of the most challenging data splits, yielding an accuracy of merely 87.5%. Upon conducting a comprehensive analysis, we identified a total of 226 misclassified sequences among the 1,769 instances examined. Of these misclassifications, 158 were classified as false positives, while 68 were categorized as false negatives.

Further examination highlighted the difficulty in distinguishing between genuine contractions and cow fidgeting, even for trained experts. However, in real-world scenarios, the evaluation of contractions becomes comparatively more feasible by considering the temporal context. This is attributed to the likelihood that intense contractions are typically followed by subsequent contractions, emphasizing the significance of predicting trends rather than individual actions.

# Discussion

The topic of Contraction Detection in calving dairy cattle, as previously mentioned, is not as popular as cow lameness detection or cattle pose estimation. Finding similar works in the literature is quite challenging.

Although we may anticipate a comparable approach to our work in terms of neural network selection [23,24] there might be differences in terminology. Our method is associated with Action Recognition, which is primarily understood in the Computer Vision community as the recognition of activities in videos and requires temporal information. However, the mentioned works employ an object detection strategy to address action recognition. If we consider a video-based approach that is comparable to ours from a computer science perspective, we can refer to works [25, 26]. Nevertheless, these articles pertain to different animal species, namely horses and pigs. They both utilize CNN-LSTM models, which are also video-based. However, they belong to a different family of neural models compared to our choices. Importantly, these works do not address the issue of giving birth from a zootechnical or veterinary perspective. Instead, the researchers focus on recognizing pain in horses and action recognition of tail-biting behaviour in pigs.

The only found article partially devoted to cow births is [21]. Authors show that computer vision can be successfully applied to predict individual dairy cow behaviours with an accuracy of around 80%. Here there is a connection to our work because among studied behaviours like 'stand', 'eating', and 'drinking' there are 'contractions'. In the related work, we can read: 'the detection of birth contractions was correctly predicted 83% of the time, which in itself can be an early warning calving alert, as all cows start contractions several hours prior to giving birth'. We do not fully agree, because the time between contractions start and giving birth strongly depends on the individual cow and whether she has given birth before. Nevertheless, it is still the approach that we follow. Another similarity is the used data which is video that causes the selection of a specific class of neural models that require video clips as input. Authors used here Non-local Neural Networks [22].

Despite a similar approach, we need to stress that our work is fully concentrated on contractions recognition, and this brings a few major differences. First of all, in [21] authors captured 613 cow contractions actions while our dataset consists of 8500. It is difficult to say how many cows showed contractions in the described work. The authors claim that their research includes 46 cows that were monitored continuously under 24 h video surveillance. Taking all of this into account we assume that around 10 cows showed contractions which seems a bit too little to make valid claims related to the robustness of the adopted solution. In our case, we talk about 100 recordings of 100 different cows that end with cow birth.

Results reported in [21] are related to the previous paragraph. If for validation there are used 112 contraction actions then it is difficult to compare to our results that are characterized by 10 times bigger size of dataset. The other issue is related to the methodology. Thanks to our approach of excluding one box from the training dataset we can be sure that there is no data leakage related to the specific cow. We do not know details of an evaluation protocol used in [21], however, given the alleged number of pregnant cows that showed contractions and total number of them, we see that it is difficult to design a robust methodology. For these reasons, we claim that comparing accuracy values between works may not be valid, because results stated in [21] might be overestimated. Nevertheless, still all our models are characterized by higher values of accuracy, from 2% to nearly 10%.

The reasons behind the remarkable attainment of high accuracy values in this study can be ascribed to several key factors. Firstly, the utilization of a large-scale dataset encompassed a wide array of instances and scenarios, facilitating a comprehensive understanding of the data and enabling the model to learn from diverse examples. Variety of examples was empowered by the incorporation of augmentation techniques that enhanced the model's ability to generalize.

Furthermore, the integration of advanced architectures played a crucial role in achieving high accuracy values. For example, better results of Multiscale Vision Transformer are connected with facilitating better representation and extraction of more meaningful features from the data. Lastly, hyperparameter optimization served as a vital component in fine-tuning the models' performance.

Satisfied with the results we must be aware of a few issues. While doing everything we can in terms of the quality of the data set, there may still be more activities that we are not aware of that could be misclassified as contractions. In future there will be a necessity to perform rigorous tests of our system on recordings obtained in customer barns and eventually create an enriched dataset that consists of both contractions and very similar to them activities. This will allow us to re-train our neural models and make our solution even more robust. Also possesing new videos from customers will lead to increased variety of cows breeds, lightning conditions and, despite our camera setup recommendations, different viewpoints forced by barns architecture.

To sum up, the integration of deep learning technologies into the dairy industry holds tremendous potential for enhancing productivity, improving animal welfare, and minimizing economic losses. By providing real-time insights and early detection of complications during calving processes, such a system can enable timely interventions, reducing the risk of adverse outcomes and promoting the well-being of both cows and their offspring. The proposed solution also contributes to automation and allows farmers to use their time more effectively.

# Conclusion

In this study, we focused on the automatic recognition of cow contractions using computer vision techniques. We developed a specialized Contraction dataset comprising thousands of cow contraction actions and non-contraction sequences, recorded from a diverse set of cameras in a calving hall. The dataset allowed us to train and evaluate our models effectively, taking into account variations in camera views and cow behavior. Finally,we setup the end-to-end system in the high-input farm to evaluate its working.

The main conclusions are as follows.

1. Both models, namely Slow R50 and Multiscale Vision Transformer (MViT) achieved high results, depending on the input variant the results range from 86.96% to 92.72% and there is a tradeoff between accuracy (higher for MViT)  and speed (faster for Slow R50).
2. Slow R50 appeared to be enough to get satisfactionary results, so we were able to ommit Fast path from the model achieving higher speed results in both training and inference phases
3. Righours tests in the high-input farm revealed necessity of acquiring even bigger dataset (currently one hundred births of cows with annotated contractions) and diagnosing activities that cause false positives and furher including them in the training datasets to make neural models more robust.

In conclusion, our study demonstrates the feasibility of utilizing computer vision techniques for cow contraction recognition. The developed models, along with the comprehensive Contraction dataset, contribute to the advancement of automated analysis and monitoring systems in the dairy industry, ultimately enhancing reproductive management and animal welfare.

**Bibliography**

1. Barrier AC, Ruelle E, Haskell MJ, Dwyer CM. Effect of a difficult calving on the vigour of the calf, the onset of maternal behaviour, and some behavioural indicators of pain in the dam. Preventive Veterinary Medicine. 2012 Mar 1;103(4):248-56.
2. Barrier AC, Haskell MJ, Birch S, Bagnall A, Bell DJ, Dickinson J, Macrae AI, Dwyer CM. The impact of dystocia on dairy calf health, welfare, performance and survival. The Veterinary Journal. 2013 Jan 1;195(1):86-90.
3. Lombard JE, Garry FB, Tomlinson SM, Garber LP. Impacts of dystocia on health and survival of dairy calves. Journal of dairy science. 2007 Apr 1;90(4):1751-60.
4. Meyer CL, Berger PJ, Koehler KJ, Thompson JR, Sattler CG. Phenotypic trends in incidence of stillbirth for Holsteins in the United States. Journal of Dairy Science. 2001 Feb 1;84(2):515-23.
5. Schuenemann GM, Nieto I, Bas S, Galvão KN, Workman J. Assessment of calving progress and reference times for obstetric intervention during dystocia in Holstein dairy cows. Journal of dairy science. 2011 Nov 1;94(11):5494-501.
6. Miedema HM, Cockram MS, Dwyer CM, Macrae AI. Changes in the behaviour of dairy cows during the 24 h before normal calving compared with behaviour during late pregnancy. Applied Animal Behaviour Science. 2011 Apr 1;131(1-2):8-14.
7. Sheldon IM, Cronin J, Goetze L, Donofrio G, Schuberth HJ. Defining postpartum uterine disease and the mechanisms of infection and immunity in the female reproductive tract in cattle. Biology of reproduction. 2009 Dec 1;81(6):1025-32.

8. Wehrend A, Hofmann E, Failing K, Bostedt H. Behaviour during the first stage of labour in cattle: Influence of parity and dystocia. Applied Animal Behaviour Science. 2006 Nov 1;100(3-4):164-70.
9. Schirmann K, Chapinal N, Weary DM, Vickers L and von Keyserlingk MAG, Rumination and feeding behavior before and after calving in dairy cows. J Dairy Sci 96:7088–7092 (2013).
10. Manjunath BS, Salembier P, Sikora T, editors. Introduction to MPEG-7: multimedia content description interface. John Wiley & Sons; 2002 Jun 14.
11. Khan A, Sohail A, Zahoora U, Qureshi AS. A survey of the recent architectures of deep convolutional neural networks. Artificial intelligence review. 2020 Dec;53:5455-516.
12. Dosovitskiy A, Beyer L, Kolesnikov A, Weissenborn D, Zhai X, Unterthiner T, Dehghani M, Minderer M, Heigold G, Gelly S, Uszkoreit J. An image is worth 16x16 words: Transformers for image recognition at scale. arXiv preprint arXiv:2010.11929. 2020 Oct 22.
13. Kang X, Zhang XD, Liu G. A review: development of computer vision-based lameness detection for dairy cows and discussion of the practical applications. Sensors. 2021 Jan 22;21(3):753.
14. Bezsonov O, Lebediev O, Lebediev V, Megel Y, Prochukhan D, Rudenko O. Breed recognition and estimation of live weight of cattle based on methods of machine learning and computer vision. Eastern-European Journal of Enterprise Technologies. 2021 Dec 29;6(9):114.
15. Nguyen C, Wang D, Von Richter K, Valencia P, Alvarenga FA, Bishop–Hurley G. Video-based cattle identification and action recognition. In2021 Digital Image Computing: Techniques and Applications (DICTA) 2021 Nov 29 (pp. 01-05). IEEE.
16. Manoharan DS. Embedded imaging system based behavior analysis of dairy cow. Journal of Electronics and Informatics. 2020 Jun 6;2(2):148-54.
17. Liseune A, Salamone M, Van den Poel D, Van Ranst B, Hostens M. Predicting the milk yield curve of dairy cows in the subsequent lactation period using deep learning. Computers and Electronics in Agriculture. 2021 Jan 1;180:105904.
18. Li X, Cai C, Zhang R, Ju L, He J. Deep cascaded convolutional models for cattle pose estimation. Computers and Electronics in Agriculture. 2019 Sep 1;164:104885.
19. Gong C, Zhang Y, Wei Y, Du X, Su L, Weng Z. Multicow pose estimation based on keypoint extraction. PloS one. 2022 Jun 3;17(6):e0269259.
20. Fan H, Xiong B, Mangalam K, Li Y, Yan Z, Malik J, Feichtenhofer C. Multiscale vision transformers. InProceedings of the IEEE/CVF international conference on computer vision 2021 (pp. 6824-6835).
21. McDonagh J, Tzimiropoulos G, Slinger KR, Huggett ZJ, Down PM, Bell MJ. Detecting dairy cow behavior using vision technology. Agriculture. 2021 Jul 17;11(7):675.
22. Wang X, Girshick R, Gupta A, He K. Non-local neural networks. InProceedings of the IEEE conference on computer vision and pattern recognition 2018 (pp. 7794-7803).
23. Nguyen C, Wang D, Von Richter K, Valencia P, Alvarenga FA, Bishop–Hurley G. Video-based cattle identification and action recognition. In2021 Digital Image Computing: Techniques and Applications (DICTA) 2021 Nov 29 (pp. 01-05). IEEE.
24. Andersen PH, Broomé S, Rashid M, Lundblad J, Ask K, Li Z, Hernlund E, Rhodin M, Kjellström H. Towards machine recognition of facial expressions of pain in horses. Animals. 2021 Jun 1;11(6):1643.
25. Liu D, Oczak M, Maschat K, Baumgartner J, Pletzer B, He D, Norton T. A computer vision-based method for spatial-temporal action recognition of tail-biting behaviour in group-housed pigs. Biosystems Engineering. 2020 Jul 1;195:27-41.
26. Boris Sekachev, Nikita Manovich, Maxim Zhiltsov, Andrey Zhavoronkov, Dmitry Kalinin, Ben Hoff, TOsmanov, Dmitry Kruchinin, Artyom Zankevich, DmitriySidnev, Maksim Markelov, Johannes222, Mathis Chenuet, a-andre, telenachos, Aleksandr Melnikov, Jijoong Kim, Liron Ilouz, Nikita Glazov, … Tritin Truong. (2020). opencv/cvat: v1.1.0 (v1.1.0). Zenodo. https://doi.org/10.5281/zenodo.4009388